\documentclass{article}

\usepackage[preprint]{neurips_2025}

\usepackage[utf8]{inputenc} 
\usepackage[T1]{fontenc}    
\usepackage{hyperref}       
\usepackage{url}            
\usepackage{booktabs}       
\usepackage{amsfonts}       
\usepackage{nicefrac}       
\usepackage{microtype}      
\usepackage{xcolor}         
\usepackage{amsmath}
\usepackage{graphicx}
\usepackage{subcaption}

\newcommand{\M}{\mathcal{M}}
\newcommand{\F}{\mathcal{F}}
\newcommand{\K}{\mathcal{K}}
\newcommand{\A}{\mathcal{A}}
\newcommand{\Enc}{\mathsf{Enc}}
\newcommand{\Dec}{\mathsf{Dec}}

\title{Can Transformers Break Encryption Schemes via In-Context Learning?}

\author{%
Jathin Korrapati\thanks{Equal contribution. Author order randomized.} \quad
Patrick Mendoza$^{*}$ \quad
Aditya Tomar$^{*}$ \quad
Abein Abraham$^{*}$ \\[0.5em]
UC Berkeley \\
\texttt{\{jkorr, patmendoza6745, adityatomar, aabraham\}@berkeley.edu}
}

\begin{document}

\maketitle

\begin{abstract}
In-context learning (ICL) has emerged as a powerful capability of transformer-based language models, enabling them to perform tasks by conditioning on a small number of examples presented at inference time, without any parameter updates. Prior work has shown that transformers can generalize over simple function classes like linear functions, decision trees, even neural networks, purely from context, focusing on numerical or symbolic reasoning over underlying well-structured functions. Instead, we propose a novel application of ICL into the domain of cryptographic function learning, specifically focusing on ciphers such as mono-alphabetic substitution and Vigenère ciphers, two classes of private-key encryption schemes. These ciphers involve a fixed but hidden bijective mapping between plain text and cipher text characters. Given a small set of (cipher text, plain text) pairs, the goal is for the model to infer the underlying substitution and decode a new cipher text word. This setting poses a structured inference challenge, which is well-suited for evaluating the inductive biases and generalization capabilities of transformers under the ICL paradigm. Code is available at \url{https://github.com/adistomar/CS182-project}.
\end{abstract}

\section{Introduction}
Large language models such as GPT-3 \cite{brown2020language} have demonstrated an impressive ability to perform in-context learning: given a prompt containing examples of a task (input-output pairs) and a new query input, the model can generate the appropriate output without additional parameter updates. This intriguing capability raises the question of whether such models can learn entirely new tasks on the fly, or whether they are merely retrieving memorized patterns from pretraining \citep{brown2020language, J1WhitePaper, rae2022scalinglanguagemodelsmethods, gpt-neox-20b}. 

A landmark study by ~\cite{garg2022can} formalized this question by proposing a precise framework for in-context learning: they analyzed whether Transformer models can be explicitly trained to learn entire \textit{function classes} from examples presented in the prompt. Rather than just recognizing familiar tasks, their work showed that Transformers can act as meta-learners, adapting to unseen functions using only in-context examples drawn from known function families. Building on this framework, we investigate whether Transformer models can in-context learn structured tasks drawn from \textit{private-key encryption schemes}. 

Formally, a private-key encryption scheme $\mathcal{S}$ is defined by a message space $\M$, a key space $\K$, an encryption function $\Enc(k, m) = \Enc_{k}(m)$, and a decryption function $\Dec(k, c) = \Dec_{k}(c)$. The correctness condition guarantees that decrypting an encrypted message yields the original plaintext: $\Dec_{k}(\Enc_{k}(m)) = m$ for all $m \in M$, $k \in K$. For a given scheme $S$, we can define the class of encryption functions as $\mathcal{F}_S = \{\Enc_k : k \in K\}$. A prompt $P$ is generated by first sampling a message $m$ from the message space $\M$ drawn randomly according to the distribution \( D_{\mathcal{M}} \). We then compute its encryption \( c = \Enc_k(m) \) and construct the prompt \( P \) as a sequence:
\[
P = \left( c[1], m[1], c[2], m[2], \ldots, c[l], m[l], c[l + 1] \right),
\]
where we pair each ciphertext character \( c[j] = \Enc_k(m[j]) \) with its corresponding plaintext character \( m[j] \) for \( j = 1, \ldots, l \) while only revealing the ciphertext \( c[l + 1] \) of the next character, without revealing the plaintext \( m[l + 1] \). The goal is to use a model $M$ to predict $\Dec_k(c[l + 1])$.

We empirically study this question by exploring whether Transformer models can in-context learn across a \textit{range of private-key encryption schemes}. Specifically, we investigate whether models can generalize decryption behavior for previously unseen keys when provided with ciphertext-plaintext pairs as in-context examples. As a concrete example, we examine the \textit{mono-alphabetic substitution cipher}, a classical scheme where each letter in the alphabet is permuted. More broadly, we extend our analysis to additional encryption schemes, such as the \textit{Vigenère cipher}, to assess the generality and limits of in-context learning in cryptographic settings ~\cite{katz2014modern}.

To benchmark performance, we compare the model’s predictions against established cryptanalytic attacks tailored to each scheme, such as frequency analysis for substitution ciphers. By systematically evaluating across multiple encryption methods, we aim to understand whether Transformer models can \textit{meta-learn general decryption strategies} from in-context examples, extending the insights of ~\cite{garg2022can} on function class learning into the domain of cryptography.

\section{Related Work}
\label{sec:related}
\paragraph{In-context learning and algorithmic reasoning.}
Large transformer models such as GPT-3 have shown that next-token prediction alone can induce powerful
\emph{in-context learning} (ICL) behaviours, enabling the model to fit new mappings from just a handful of
demonstrations \citep{brown2020language}.
These models demonstrate an ability to adapt to novel tasks presented through examples within the input prompt, achieving this adaptation without requiring updates to their parameters \citep{aitchison2023bayesianmeta, kirsch2022general, garg2022can}. This phenomenon allows models to perform what appears to be few-shot learning during inference, even potentially addressing tasks not encountered during their pre-training phase \citep{chen2024bypassing, wang2025can}. Some studies posit that the Transformer architecture, particularly its attention mechanism, simulates optimization algorithms or learning processes during its forward pass \citep{chen2024bypassing, kirsch2022general} Foundational work by \cite{garg2022can} established that Transformers, when trained appropriately on specific function classes like linear regression, can achieve performance comparable to classical estimators such as the optimal least squares solution, learning effectively from in-context examples. Others have also proposed that Transformers can perform full Bayesian inference in context for certain statistical models \citep{reuter2025can}. Indeed, research has explored explicitly meta-training Transformers to function as general-purpose in-context learners, observing transitions between memorization and generalization behaviors influenced by factors like model size and the diversity of meta-training tasks \citep{kirsch2022general}. Architectures like Looped Transformers, which employ weight sharing across layers, have been theoretically and empirically shown to implement multi-step GD or iterative algorithms such as Chebyshev iteration, potentially offering improved out-of-distribution (OOD) generalization compared to standard deep Transformers \citep{chen2024bypassing, yang2024looped}.

\paragraph{Neural approaches to cipher decipherment.}
Applying deep networks to cryptanalysis predates modern LLMs.
\citet{greydanus2017neuralcipher} used an LSTM to learn Enigma-style polyalphabetic ciphers from
supervised data and \citet{dani2024breaking} used multiple deep learning architectures to comprehensively evaluate cryptographic indistinguishability of lightweight block ciphers, which included LSTMs. Additionally, \citet{gomez2018ciphergan} introduced \textsc{CipherGAN}, an unsupervised
GAN that recovered mono-alphabetic and Vigenère keys by aligning plaintext and ciphertext
distributions. Building on this, the Unified Cipher GAN (UC-GAN) was proposed as a single model capable of performing ciphertext-to-plaintext (and vice-versa) translations across multiple classical substitution cipher types (treated as different domains), achieving high accuracy without prior knowledge of the ciphers \citep{parkcryptography7030035}. More contemporary work uses supervised learning with ResNets to challenge the indistinguishability property of modern block ciphers like SPECK32/64 under Known Plaintext Attack (KPA) settings, training models to distinguish ciphertexts originating from plaintexts that differ by only a single bit \citep{dani2024breaking}. Supervised binary classification models have also been developed to directly model Indistinguishability under Chosen Plaintext Attack (IND-CPA) security games, learning to differentiate ciphertexts produced from specific, chosen plaintexts (e.g., all-zero plaintexts versus uniformly random plaintexts) \citep{kim2025cryptanalysis}. Furthermore, Transformers have assisted lattice-based attacks \citep{wenger2023salsa} and distinguished block-cipher outputs
\citep{gohr2019speck}.

\paragraph{LLMs as few-shot codebreakers} An orthogonal line of research investigates the capabilities of large, pre-trained language models (LLMs)---such as GPT-3, GPT-4, Claude, and Llama---when applied directly to classical cipher decipherment tasks. Unlike the previously discussed neural cryptanalysis approaches that often involve specialized architectures or task-specific training, this research leverages the models' existing, general-purpose pre-training and their emergent in-context learning abilities. These models are typically prompted in a zero-shot or few-shot manner, provided with the ciphertext and perhaps a small number of plaintext-ciphertext examples, but without any further fine-tuning on cryptographic data \citep{li2025cipherbank}.
Initial explorations and subsequent systematic benchmarking have revealed a consistent pattern in LLM performance on classical ciphers. Highly capable models like GPT-4 demonstrate a surprising ability to correctly decrypt a substantial fraction of simple classical ciphers, particularly substitution ciphers like Caesar, Rot13, or Atbash, when prompted with few or even zero examples. However, this success does not extend to ciphers that involve more complex structural manipulations, polyalphabetic keys (e.g., Vigenère), transposition, or algorithms requiring multiple steps of logical deduction \citep{li2025cipherbank}. Performance drops dramatically on such ciphers, mirroring the limitations observed in the broader ICL literature concerning compositional reasoning and algorithmic complexity.

\paragraph{Our contribution.}
Unlike prior neural decipherment systems that \emph{train} directly on cipher text
\citep{aldarrab2021sequencetosequencemodelscracksubstitution,kamb2023decipher}, we investigate a Transformer's ability to
\emph{learn a private-key encryption scheme on the fly} via ICL, bridging meta-learning studies
\citep{garg2022can} with neural cryptanalysis.  By supplying only a few $(\text{cipher},\text{plain})$
pairs at inference time, we test the limits of transformers as few-shot codebreakers and analyze how
this structured task differs from previously studied numeric function classes.

This paper builds directly on the work of \cite{garg2022can}, which demonstrated that transformers are capable of performing in-context learning (ICL) tasks using only example sequences provided at inference time. That earlier study focused on simple function classes---such as linear functions, decision trees, and basic MLPs---showing that transformers can effectively emulate classical learning algorithms by pattern-matching over examples. Subsequent research has expanded this exploration to broader function classes, examined the limits of transformer capabilities, and investigated the connection between ICL and meta-learning.

Our project advances this line of inquiry by shifting attention from learning general linear functions to tackling cryptographic functions. Specifically, instead of learning a linear mapping from $(x, f(x))$ pairs, we challenge the transformer to learn the decryption function $\Dec_k(\cdot)$ from (ciphertext, plaintext) pairs. This reframes ICL through a new lens on how transformers can be leveraged for learning tasks in the domain of cryptography.

\section{Method}
We now describe our methodology for training a model that can in-context learn a function class $\F_{\mathcal{S}}$ with respect to a distribution $D_\mathcal{F_S}$ over functions, and $D_{\mathcal{M}}$ over messages. To generate a single training prompt, we first sample a random encryption function $\Enc_k$ from the class according to $D_{\F_{\mathcal{S}}}$, then sample a message $m$ according to $D_{\M}$, and finally evaluate $\Enc_k$ on these inputs to produce the prompt $P = \left( c[1], m[1], c[2], m[2], \ldots, c[l], m[l], c[l + 1] \right)$. For example, in the case of the mono-alphabetic substitution cipher, where the alphabet $\mathcal{A}$ consists of the lowercase letters (a-z), and $\M$ is a language modeling dataset composed of lowercase English sentences, containing no punctuation, numbers, special characters, or spaces, inputs could be uniformly drawn from $\M$, and a random Encryption function chosen by sampling a key $k \sim U\left(\{0, \ldots, |\mathcal{A}|\}\right)$ to obtain $\Enc_k$ and $\Dec_k$. 

Given this prompt format, we train a model to predict $m[j]$ for a given $c[j]$ based on a set of preceding in-context examples. Concretely, let $P^j$ denote the prompt prefix containing $j$ in-context examples (the first $j$ ciphertext-plaintext pairs) and the $(j +1)^{\text{th}}$ input:
$P^j = (c[1], m[1], c[2], m[2], \ldots, c[j], m[j], c[j + 1])$. Then, we train a model $M_\theta$ parametrized by $\theta$ aiming to minimize the cross-entropy loss over all the prompt prefixes:
\begin{equation}
\label{eq:objective}
\min_\theta \; \mathbb{E}_P \left[ \frac{1}{l+1} \sum_{j=0}^{l} \mathcal{L}\left(M_\theta(P^j), m[j+1]\right) \right]
\end{equation}
Note that the prediction for the plaintext character $m[j + 1]$ is conditioned on all input ciphertext characters $c[1], c[2], \ldots, c[j]$ and corresponding plaintext characters $m[1], m[2], \ldots, m[j]$. We apply a causal language modeling mask, similar to that used in standard LLM training, within our own objective. However, we exclude from the loss any predictions the model makes for plaintext tokens $c[j]$, since our goal is solely to learn the mapping from ciphertext to plaintext.

\paragraph{Model Structure} Similar to that of \cite{garg2022can}, we use a decoder-only transformer model based on the GPT-2 architecture (\cite{radford2019language}) consisting of 12 layers, 8 attention heads, and a 256-dimensional embedding space (9.5M parameters). This architecture takes as input a sequence of vectors from its embedding space and predicts the next vector in that same space (in the case of language modeling, these vectors represent input tokens). We map each prompt inputs $c[l + 1] = \Enc_k(m[l + 1])$ to the same dimension as prompt outputs $m[l + 1]$. Note that the Transformer architecture allows us to compute the prediction $M_\theta(P^j)$ for all prompt prefixes in a single forward pass.

\paragraph{Tokenization.} Regarding tokenization, we throw away any default tokenizer for the model and instead use a simple vocabulary consisting of only 26 tokens consisting of the lower case alphabet (a-z) in order. We thus redefine the model's embedding matrix to 26 vectors of dimension 256, corresponding to the 26 tokens.

\paragraph{Training.} We train the model according to the training objective ~\eqref{eq:objective}. We do so by sampling a batch of random prompts at each training step and then updating the model through a gradient update. We defer to Appendix~\ref{sec:trainingdetails} for specific training details. This training is done from scratch, that is, we do not fine-tune a pre-trained language model.

\section{In-Context Learning of Mono-Alphabetic Substitution Ciphers}
We first begin with the \textbf{mono-alphabetic substitution cipher}, where each letter in the alphabet is mapped to another letter, and then applied to the plaintext.

Formally, let $\A$ be the set of lower case English alphabet letters in our message space $\M$ (all strings of finite length), and let $\mathcal{K}=\{\text{all bijections/permutations of the alphabet } \pi: \A \to \A\ \}$, so $k=\pi \in \K$ is some permutation of the set of characters in our message space. We can then define encryption and decryption as follows:

\begin{align*}
\Enc_k(m) &= \Enc_k(m[1], m[2], \ldots, m[l]) = (c[1], c[2], \ldots, c[l]), \quad \text{where } c[i] = \pi(m[j]) \\
\Dec_k(c) &= \Dec_k(c[1], c[2], \ldots, c[l]) = (m[1], m[2], \ldots, m[l]), \quad \text{where } m[j] = \pi^{-1}(c[j])
\end{align*}

where $k=\pi$ is uniformly drawn from $\K$.  We can implement this encryption function to generate prompts for training and inference as described above. 

\paragraph{Message distribution.} For sampling messages \( m \in \mathcal{M} \), we use the OpenWebText (\cite{Gokaslan2019OpenWebText}) corpus, preprocessing it by lowercasing and removing punctuation, digits, and spaces to obtain sequences over the alphabet $\A$. We do this to ensure that the training data matches the typical distribution of English letters. 

\paragraph{Baselines. } To evaluate the performance of our model on mono-alphabetic substitution, we compare it against two decoding strategies: (a) a naive lookup decoder, which stores (cipher, plain) pairs observed in the prompt and returns a space for unseen ciphertexts, and (b) a frequency-based decoder, which fills in unseen ciphertext characters with the next most common unused English letter. The naive decoder provides a strict upper bound on any exact-match model's performance given the available in-context examples, while the frequency-based decoder simulates a classical cryptanalytic heuristic. These serve as interpretable baselines to evaluate whether the transformer learns beyond memorization or frequency matching.

\subsection{Transformers can in-context learn mono-alphabetic substitution ciphers}

\begin{figure}
    \centering
    \includegraphics[width=1\linewidth]{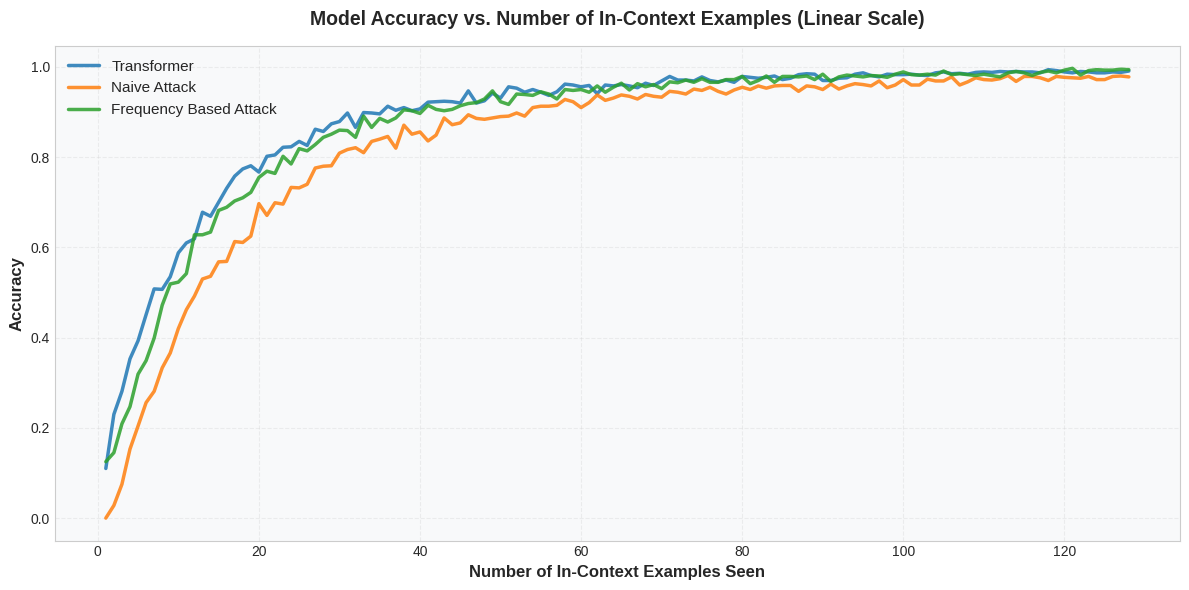}
    \caption{Evaluating the trained Transformer on in-context learning mono-alphabetic substitution ciphers. The Transformer's accuracy increases at a similar rate to both evaluation attacks, and seems to be slightly better than both.}
    \label{fig:mono-alphabetic-in-distribution}
\end{figure}

We show the in-context learning ability of the trained Transformer on the mono-alphabetic substitution cipher in Figure~\ref{fig:mono-alphabetic-in-distribution}, alongside two baseline decoders. The Transformer consistently outperforms both baselines as the number of in-context examples increases, achieving high accuracy even when the test ciphertext character was not seen verbatim in the prompt. This suggests the model is not merely copying but learning the underlying distribution of English letters, which is how it can achieve accuracy similar to and slightly better than that of the frequency based attack. 

We argue that memorization of training prompts cannot explain this performance. The key space of mono-alphabetic substitution is the set of all permutations over 26 characters, which contains $26! \approx2^{88}$
  unique keys. This is large enough to ensure the model never sees the same key twice during training. Similarly, the message space consists of arbitrary English character sequences drawn from a large corpus (OpenWebText (\cite{Gokaslan2019OpenWebText})), with no repeated messages or prompts. The total number of unique prompts seen during training exceeds 250,000, and each prompt is generated with an independent random key. This makes direct memorization infeasible. Instead, we find that the model learns to use the alternating (cipher, plain) pattern in the prompt to build an internal representation of the key, and apply it to unseen ciphertext characters during inference, capturing a structured in-context learning algorithm for mono-alphabetic cipher decryption.

\subsection{Extrapolating beyond the training distribution}
To test whether the model’s in-context learning capabilities extend beyond the specific data distribution it was trained on, we evaluate it on inputs that deviate significantly from natural language. In particular, we generate test prompts by sampling sequences of characters uniformly at random from the alphabet $A$, rather than drawing from English text as in training. These synthetic messages contain no linguistic structure or frequency biases, thus removing any cues the model may have implicitly relied on during training.

\begin{figure}
    \centering
    \includegraphics[width=1\linewidth]{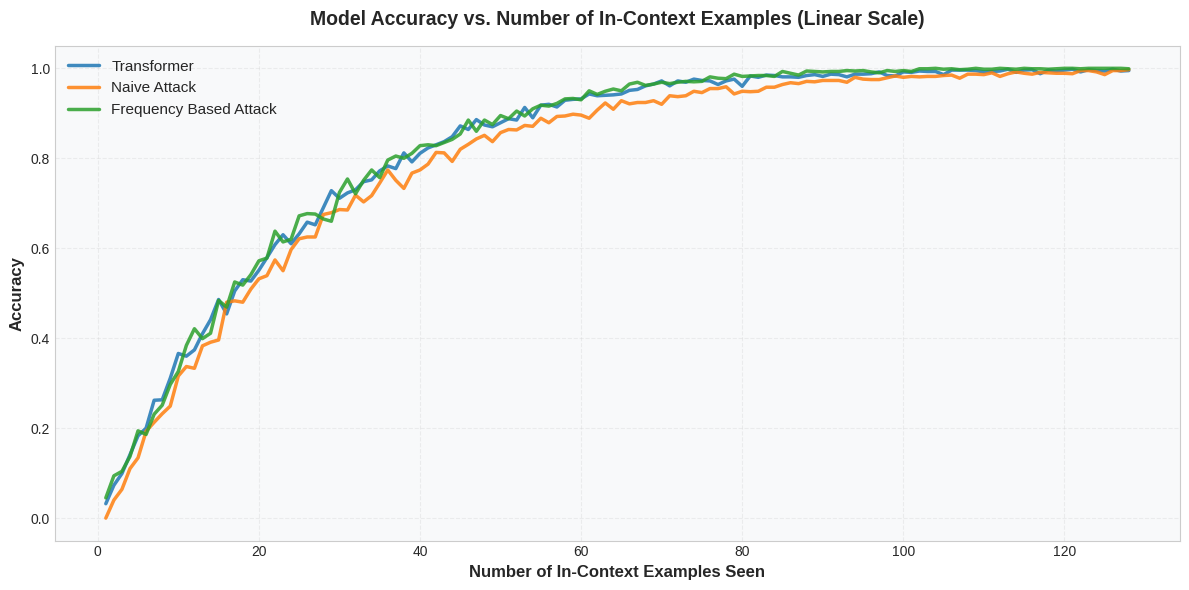}
    \caption{In-context learning of out of distribution prompts. The model generalizes well when we sample uniformly random letters instead of standard English text.}
    \label{fig:mono-alphabetic-ood}
\end{figure}

Despite this distributional shift, the Transformer still performs remarkably well, as shown in Figure~\ref{fig:mono-alphabetic-ood}. Its decryption accuracy on uniformly sampled test sequences is still very close to those of the known attacks. This result highlights a key strength of in-context learning: the model has not merely memorized statistical properties of English text, but has instead learned to infer substitution rules based on the pattern of alternating ciphertext and plaintext pairs. The fact that performance remains high on out-of-distribution inputs suggests that the model has internalized a general algorithm for mono-alphabetic decryption, rather than overfitting to distribution of the training data. However, as expected, to reach the same accuracy the model requires more in-context examples for out-of-distribution tasks (such as random prompts) than it does for in-distribution ones.
  
\section{In-Context Learning of Fixed-Length Vigenère Ciphers}

The \textbf{Vigenère cipher} is a polyalphabetic substitution cipher that uses a keyword to determine a sequence of Caesar shifts for each letter in the plaintext. Unlike the mono-alphabetic substitution cipher which uses a single fixed permutation, the Vigenère cipher varies the substitution per character based on the corresponding character of the repeating keyword.

Let $\A = \{a, b, \ldots, z\}$ be the alphabet of size 26, and let messages $m \in \M$ be strings of characters over $\A$. Let $k = k[1]\,k[2]\,\ldots\,k[\ell]$ be a keyword of length $\ell$, where each $k[j] \in \A$, and the keyword is repeated to match the length of the message. Each character in the keyword corresponds to an integer shift between 0 and 25.

Define the shift function:
\[
\text{shift}(x) = \text{ord}(x) - \text{ord}(\text{``}A\text{''})
\]
where $\text{ord}(x)$ gives the ASCII value of character $x$.

Then encryption and decryption are defined as follows:
\begin{align*}
\Enc_k(m) &= c \quad \text{where } c[j] = (m[j] + \text{shift}(k[i \bmod \ell])) \bmod 26 \\
\Dec_k(c) &= m \quad \text{where } m[j] = (c[j] - \text{shift}(k[i \bmod \ell])) \bmod 26
\end{align*}

Each letter of the plaintext is shifted forward in the alphabet by the value corresponding to the matching keyword character (repeating the keyword cyclically). Decryption shifts the letters backward using the same key.

To train and evaluate models on this cipher, we can again generate prompts using randomly sampled English plaintext messages and randomly chosen keywords. The keyspace size for the Vigenère cipher is $26^\ell$ for a keyword of length $\ell$, which grows exponentially, adding complexity over the mono-alphabetic substitution cipher.

\paragraph{Baselines.} To evaluate the model on the Vigenère cipher, we compare it to two simple decoders that assume knowledge of the key length $\ell$. (a) The naive decoder infers the key offset at each key position from in-context (cipher, plain) pairs, and uses it to decrypt matching positions; unseen positions default to a space. (b) The frequency decoder does the same but defaults to predicting 'e' when an offset is unknown, reflecting a common cryptanalytic heuristic. These baselines offer interpretable reference points for assessing whether the Transformer learns beyond memorization or frequency guessing.

\subsection{Transformers can in-context learn fixed length Vigenère ciphers}
\begin{figure}
    \centering
    \includegraphics[width=1\linewidth]{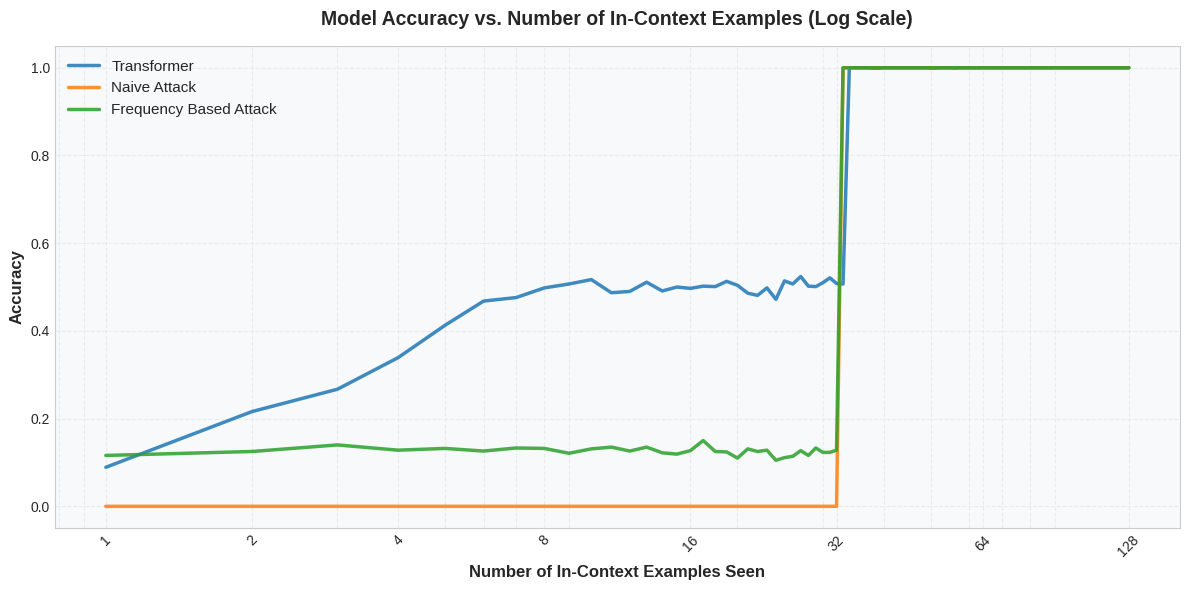}
    \caption{Evaluating the trained Transformer on in-context learning fixed-length Vigenère ciphers.}
    \label{fig:vig_32_train}
\end{figure}

We evaluate the Transformer’s ability to perform in-context learning on Vigenère ciphers with a fixed key length of 32. From Figure~\ref{fig:vig_32_train}, we see that the model achieves 100\% decryption accuracy once it has observed 32 examples, matching the performance of the known attacks. This makes sense, as after 32 examples we can easily derive the private key by observing the previous 32 shifts from plaintext to ciphertext. However, unlike these traditional approaches, the Transformer exhibits stronger generalization under partial information: when provided with fewer than 32 examples, it achieves significantly higher accuracy than the baselines, suggesting that it can partially infer or generalize key offsets even for unseen positions. 

This behavior indicates that the Transformer has learned a structured in-context decryption algorithm, one that uses positional cues, previously inferred offsets, and an understanding of the distribution of English letters to extrapolate beyond what is explicitly given. It reflects a form of algorithmic pattern completion not observed in naive frequency-based or table-based heuristics.

\subsection{Extrapolation: Sampling uniform letters}
\begin{figure}
    \centering
    \includegraphics[width=1\linewidth]{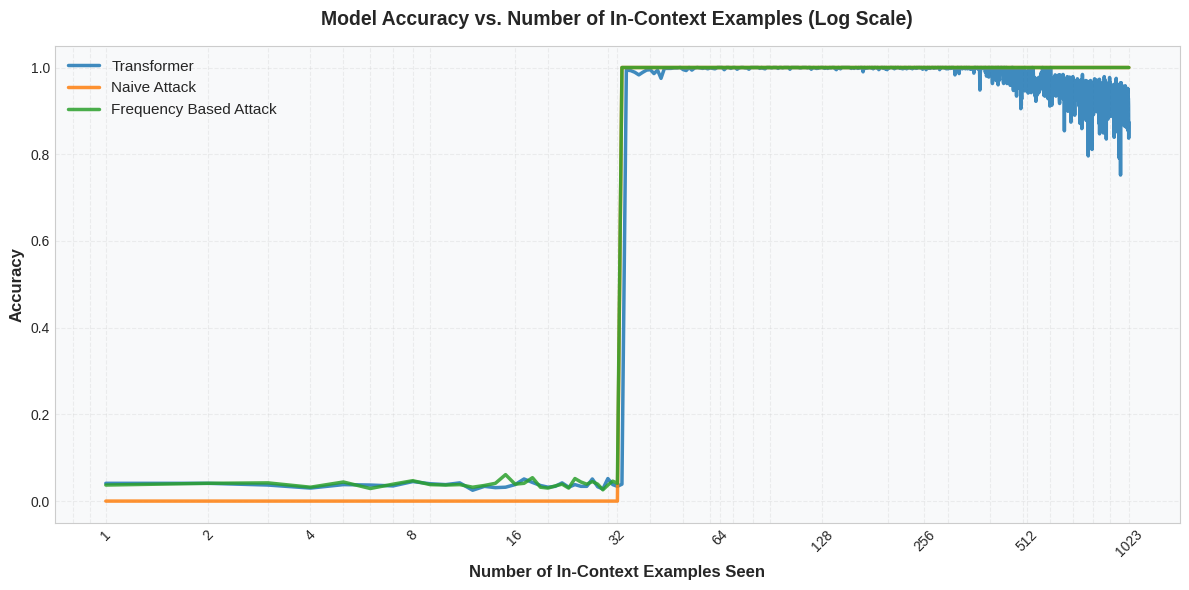}
    \caption{In-context learning of out of distribution prompts via uniform sampling for fixed-length Vigenère ciphers}
    \label{fig:vig_32_uniform_2}
\end{figure}
To assess whether the model's performance relies on language-specific patterns, we evaluate the fixed-length Vigenère model (trained on English text) on inputs where plaintext messages are sampled uniformly at random over the alphabet $A$. These inputs contain no linguistic structure or frequency biases, removing any statistical cues the model may have learned during training.

In this setting, the Transformer’s performance drops significantly when given fewer than 32 in-context examples, closely matching the accuracy of the frequency-based baseline (Figure~\ref{fig:vig_32_uniform_2}). Once 32 examples are provided, revealing all key positions, performance recovers to 100\%, as expected. This contrast highlights that the model's strong performance under partial context in the natural language setting stems largely from its implicit knowledge of English letter frequencies. When those priors are unavailable, it behaves similarly to a frequency-based decoder, indicating that generalization under partial key information is distribution-dependent.

\subsection{Extrapolation: Different key lengths at inference time}
\begin{figure}
    \centering
    \begin{subfigure}[t]{1\linewidth}
        \centering
        \includegraphics[width=0.6\linewidth]{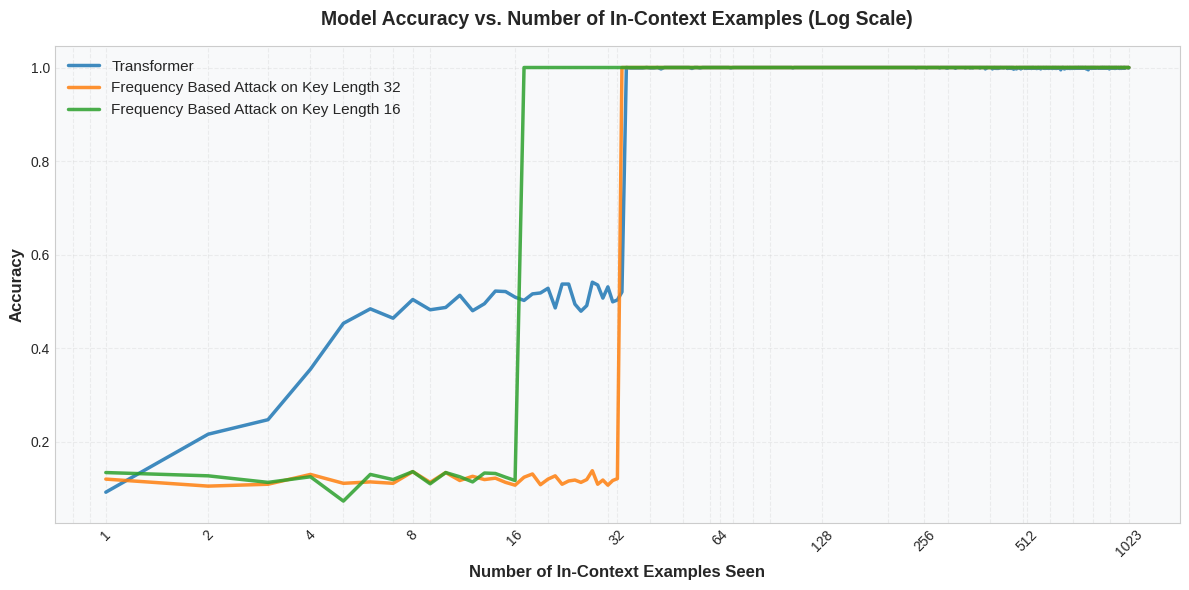}
        \caption{Key Length 16}
        \label{fig:vig_32_train_16}
    \end{subfigure}
    \hfill
    \begin{subfigure}[t]{1\linewidth}
        \centering
        \includegraphics[width=0.6\linewidth]{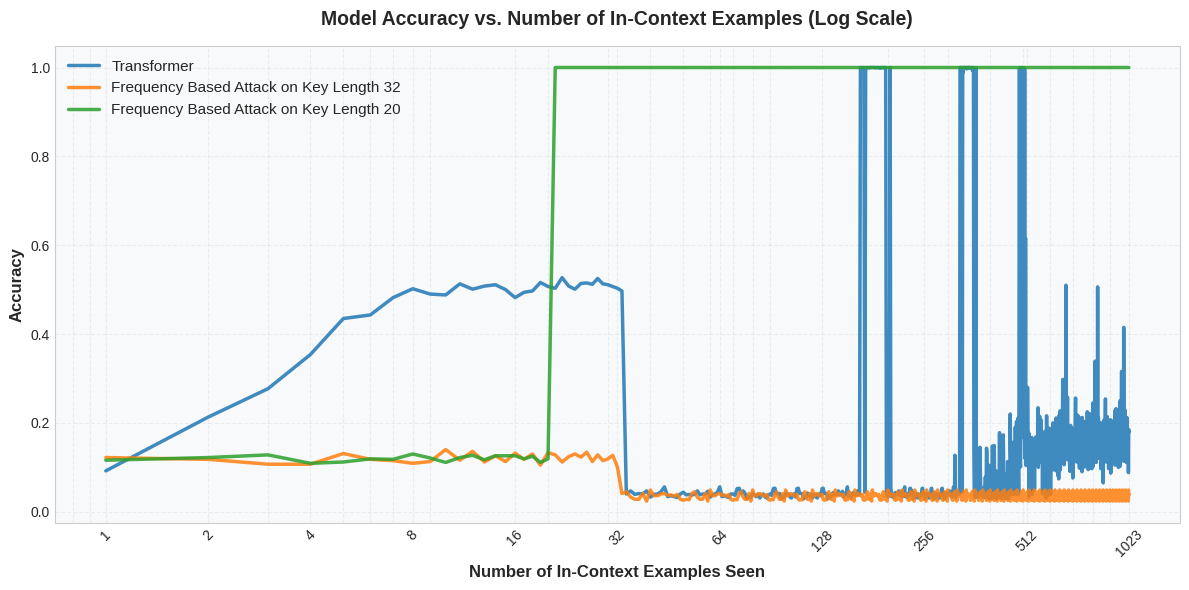}
        \caption{Key Length 20}
        \label{fig:vig_32_train_20}
    \end{subfigure}
    \caption{In-context learning out of distribution prompts via different key lengths at inference time for fixed-length Vigenère ciphers. }
    \label{fig:vig_32_train_x}
\end{figure}

To evaluate the model’s robustness to shifts in the underlying cipher structure, we test the Transformer, which was trained exclusively on Vigenère ciphers with key length 32, on prompts where the true key length differs. Specifically, we evaluate on two new prompt distributions: one where the key length is 16, and another where it is 20. The results of this are shown in Figure~\ref{fig:vig_32_train_x}. 

On the length-16 distribution, the model eventually achieves high accuracy but only after observing 32 in-context examples. This is twice the number theoretically required to fully recover the key. In contrast, a decoder specialized for key length 16 can correctly decrypt the message using just 16 examples. This suggests that the model retains an implicit inductive bias toward a fixed 32-position periodic structure and must “oversee” multiple cycles before it can correctly infer all positions.

On the length-20 distribution, the model fails entirely to learn the key. Its predictions remain near chance, showing no improvement with additional context. This indicates that the model has overfit to the 32-length structure and is unable to generalize to mismatched periodicities it has never encountered.

Together, these results demonstrate that while the Transformer can generalize to shorter-period ciphers with enough context, it lacks robustness to new key lengths and does not infer the key length itself.

\section{In-Context Learning of Variable-Length Vigenère Ciphers}

To explore whether Transformers can learn to handle ciphers with variable structure, we train a model on Vigenère ciphers where the key length is not fixed. For each training prompt, we randomly sample a key length $\ell$ uniformly from the range 4 to 32, and generate a corresponding encryption key of that length. The model must therefore learn not only to decrypt ciphertexts given a prompt, but also to implicitly infer the underlying periodicity of the key from the in-context examples alone.

This setting introduces additional complexity: the model can no longer rely on a fixed positional structure and must instead detect and adapt to different cyclic patterns across prompts. Our goal is to evaluate whether Transformers can perform such generalization over a family of encryption functions parameterized by key length, and how their in-context learning behavior compares to models trained on fixed-key variants.

\paragraph{Baseline. } To evaluate model performance in the variable-key setting, we implement a naive key-length search decoder. This baseline assumes no knowledge of the key length and attempts to infer it directly from the in-context examples. For each candidate key length 
$\ell\in[4,32]$, it maintains a separate key table and eliminates candidates whose inferred offsets become inconsistent across positions. If all valid candidates agree on the shift for a given query position, the decoder uses that value to decrypt; otherwise, it returns nothing. This approach simulates a brute-force periodicity search over possible Vigenère structures. While it performs well when the correct key length can be unambiguously recovered, it is conservative in ambiguous cases and requires many examples to resolve conflicts. It serves as a reference point for assessing whether the Transformer can learn to infer both the key content and its length implicitly from in-context examples.

\subsection{Current architecture struggles to learn variable-length Vigenère ciphers}
\begin{figure}
    \centering
    \includegraphics[width=1\linewidth]{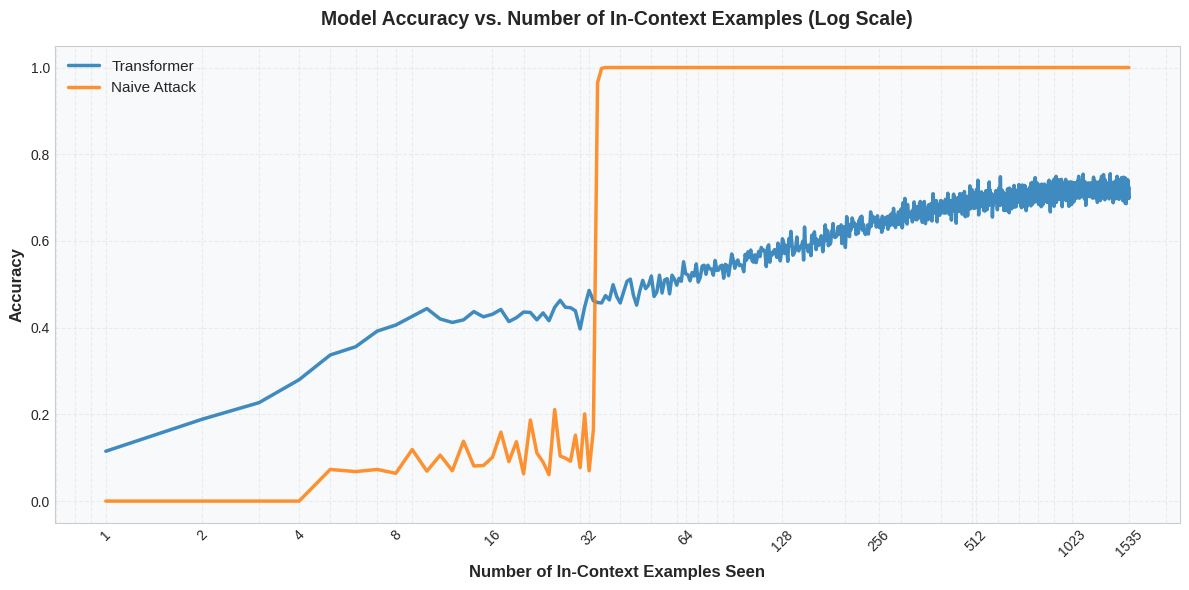}
    \caption{In-context learning for variable-length Vigenère ciphers}
    \label{fig:enter-label}
\end{figure}

Despite being trained across a range of key lengths (4-32), the model struggles to fully learn the underlying structure in the variable-key setting. As shown in Figure 6, accuracy gradually improves with the number of in-context examples but plateaus around 70\% even when provided with the maximum possible context (1535 examples in this case). In contrast, the naive key-length search baseline achieves 100\% accuracy once it observes 32 examples, which is sufficient to uniquely identify both the correct key length and all shift values. 

To improve performance in the variable-key setting, several directions remain open for exploration. First, the current model is relatively small (12-layer, 9.5M parameters), which may limit its capacity to internalize and generalize over a wide range of cipher structures. Scaling up the model, either in depth, width, or embedding size, could improve its ability to memorize periodic patterns and resolve structural ambiguity across prompts. Second, additional training may help the model better interpolate between key lengths and reinforce generalization. Curriculum learning, where training begins with fixed-length keys and gradually introduces variability, could also scaffold the model's inductive bias. 

\section{Discussion}
Our results offer strong evidence that transformer-based models, particularly in the few-shot in-context learning (ICL) setting, exhibit a non-trivial capability to learn cryptographic functions such as monoalphabetic substitution and Vigenère ciphers. While these ciphers are relatively simple from a cryptographic standpoint, they still require structured reasoning and pattern abstraction, making them ideal candidates for probing the limits of meta-learning via ICL.

We observe that transformer models successfully decipher substitution ciphers with high accuracy given a modest number of (ciphertext, plaintext) pairs in the prompt. This success underscores the model's ability to infer bijective character mappings, suggesting a strong internal inductive bias toward recognizing and generalizing symbolic relationships—even without gradient-based learning.

The Vigenère cipher, which incorporates a periodic shift based on a hidden keyword, presented a significantly harder challenge. Nonetheless, the models demonstrated partial success in recovering correct characters and identifying periodicity, especially when the keyword length was short. These results suggest that transformers can partially internalize repeated shift patterns and leverage contextual cues to hypothesize likely keys or mappings. However, success degrades rapidly with increasing key length, which mirrors the challenges faced in classical cryptanalysis and highlights current limitations in model capacity or context window.

Another important observation is that the models were able to generalize to ciphertext generated from unseen keys, reinforcing that their success stems from actual in-context generalization rather than memorization. This supports the growing view of transformers as approximate meta-learners rather than static pattern matchers. However, compared to hand-crafted cryptanalytic techniques like frequency analysis, model performance is still inconsistent and less interpretable, especially on more structured encryption schemes.

\section{Conclusion}
In this study, we investigated the capability of transformer models to perform cryptanalysis via in-context learning on classical private-key encryption schemes. Our results indicate that transformers are surprisingly competent at deciphering monoalphabetic substitution ciphers using just a few examples at inference time, revealing their capacity for symbolic reasoning and bijective mapping inference without explicit training on the task.

Extending the task to the Vigenère cipher showed that transformers can learn aspects of more complex encryption schemes, such as repeated shifts, but face challenges as structural complexity increases. These findings suggest both the promise and current limits of transformer-based ICL in cryptographic domains.

Overall, our work bridges the gap between symbolic function learning and cryptanalysis under the ICL paradigm, providing a new testbed for evaluating model inductive biases and generalization abilities. Future work may explore richer cryptographic schemes, perform more investigation on how model size or architecture impacts performance, and consider how transformers might be used as assistants in classical cryptanalytic workflows.


\bibliographystyle{plainnat}
\bibliography{references}

\begin{thebibliography}{25}
\providecommand{\natexlab}[1]{#1}
\providecommand{\url}[1]{\texttt{#1}}
\expandafter\ifx\csname urlstyle\endcsname\relax
  \providecommand{\doi}[1]{doi: #1}\else
  \providecommand{\doi}{doi: \begingroup \urlstyle{rm}\Url}\fi

\bibitem[Aitchison et~al.(2023)]{aitchison2023bayesianmeta}
Laurence Aitchison et~al.
\newblock Attention as bayesian inference: Evidence from in-context learning.
\newblock arXiv:2310.01234, 2023.
\newblock URL \url{https://arxiv.org/abs/2310.01234}.

\bibitem[Aldarrab and May(2021)]{aldarrab2021sequencetosequencemodelscracksubstitution}
Nada Aldarrab and Jonathan May.
\newblock Can sequence-to-sequence models crack substitution ciphers?, 2021.
\newblock URL \url{https://arxiv.org/abs/2012.15229}.

\bibitem[Black et~al.(2022)Black, Biderman, Hallahan, Anthony, Gao, Golding, He, Leahy, McDonell, Phang, Pieler, Prashanth, Purohit, Reynolds, Tow, Wang, and Weinbach]{gpt-neox-20b}
Sid Black, Stella Biderman, Eric Hallahan, Quentin Anthony, Leo Gao, Laurence Golding, Horace He, Connor Leahy, Kyle McDonell, Jason Phang, Michael Pieler, USVSN~Sai Prashanth, Shivanshu Purohit, Laria Reynolds, Jonathan Tow, Ben Wang, and Samuel Weinbach.
\newblock {GPT-NeoX-20B}: An open-source autoregressive language model.
\newblock In \emph{Proceedings of the ACL Workshop on Challenges \& Perspectives in Creating Large Language Models}, 2022.
\newblock URL \url{https://arxiv.org/abs/2204.06745}.

\bibitem[Brown et~al.(2020)Brown, Mann, Ryder, et~al.]{brown2020language}
Tom~B. Brown, Benjamin Mann, Nick Ryder, et~al.
\newblock Language models are few-shot learners.
\newblock \emph{Advances in Neural Information Processing Systems}, 33:\penalty0 1877--1901, 2020.
\newblock URL \url{https://arxiv.org/abs/2005.14165}.

\bibitem[Chen et~al.(2024)Chen, Li, Liang, Shi, and Song]{chen2024bypassing}
Bo~Chen, Xiaoyu Li, Yingyu Liang, Zhenmei Shi, and Zhao Song.
\newblock Bypassing the exponential dependency: Looped transformers efficiently learn in-context by multi-step gradient descent.
\newblock \emph{arXiv preprint arXiv:2410.11268}, 2024.
\newblock URL \url{https://arxiv.org/abs/2410.11268}.

\bibitem[Dani et~al.(2024)Dani, Nakka, and Saxena]{dani2024breaking}
Jimmy Dani, Kalyan Nakka, and Nitesh Saxena.
\newblock Breaking indistinguishability with transfer learning: A first look at {SPECK32/64} lightweight block ciphers.
\newblock \emph{arXiv preprint arXiv:2405.19683}, 2024.
\newblock URL \url{https://arxiv.org/abs/2405.19683}.

\bibitem[Garg et~al.(2022)Garg, Tsipras, Liang, and Valiant]{garg2022can}
Shivam Garg, Dimitris Tsipras, Percy~S Liang, and Gregory Valiant.
\newblock What can transformers learn in-context? a case study of simple function classes.
\newblock \emph{Advances in Neural Information Processing Systems}, 35:\penalty0 30583--30598, 2022.

\bibitem[Gohr(2019)]{gohr2019speck}
Annika Gohr.
\newblock Improving attacks on round-reduced speck32/64 using deep learning.
\newblock In \emph{19th IACR International Conference on Cryptology in India (Indocrypt)}, pages 284--305, 2019.

\bibitem[Gokaslan and Cohen(2019)]{Gokaslan2019OpenWebText}
Aaron Gokaslan and Vanya Cohen.
\newblock Openwebtext corpus.
\newblock \url{https://github.com/jcpeterson/openwebtext}, 2019.
\newblock Accessed: YYYY-MM-DD.

\bibitem[Greydanus(2017)]{greydanus2017neuralcipher}
Sam Greydanus.
\newblock Learning to decrypt classical ciphers with recurrent neural networks.
\newblock arXiv:1703.05775, 2017.
\newblock URL \url{https://arxiv.org/abs/1703.05775}.

\bibitem[Gómez et~al.(2018)Gómez, Li, and Fung]{gomez2018ciphergan}
Alejandro Gómez, Zhouhan Li, and Pascale Fung.
\newblock {CipherGAN}: Unsupervised cipher cracking using generative adversarial networks.
\newblock In \emph{Proceedings of the 56th Annual Meeting of the Association for Computational Linguistics}, pages 1176--1186, 2018.

\bibitem[Kambhatla et~al.(2023)Kambhatla, Born, and Sarkar]{kamb2023decipher}
Nishant Kambhatla, Logan Born, and Anoop Sarkar.
\newblock Decipherment as regression: Solving historical substitution ciphers by learning symbol recurrence relations.
\newblock pages 2136--2152, 01 2023.
\newblock \doi{10.18653/v1/2023.findings-eacl.160}.

\bibitem[Katz and Lindell(2014)]{katz2014modern}
Jonathan Katz and Yehuda Lindell.
\newblock \emph{Introduction to Modern Cryptography}.
\newblock CRC Press, Boca Raton, FL, 2nd edition, 2014.
\newblock ISBN 9781466570269.

\bibitem[Kim et~al.(2025)Kim, Vasudevan, D'Oliveira, Cohen, Stahlbuhk, and M{\'e}dard]{kim2025cryptanalysis}
Benjamin~D. Kim, Vipindev~Adat Vasudevan, Rafael G.~L. D'Oliveira, Alejandro Cohen, Thomas Stahlbuhk, and Muriel M{\'e}dard.
\newblock Cryptanalysis via machine learning based information theoretic metrics.
\newblock \emph{arXiv preprint arXiv:2501.15076}, 2025.
\newblock URL \url{https://arxiv.org/abs/2501.15076}.

\bibitem[Kirsch et~al.(2022)Kirsch, Harrison, Sohl-Dickstein, and Metz]{kirsch2022general}
Louis Kirsch, James Harrison, Jascha Sohl-Dickstein, and Luke Metz.
\newblock General-purpose in-context learning by meta-learning transformers.
\newblock \emph{arXiv preprint arXiv:2212.04458}, 2022.
\newblock URL \url{https://arxiv.org/abs/2212.04458}.

\bibitem[Li et~al.(2025)Li, Pei, Sun, Lin, Ming, Gao, Wu, He, and Wu]{li2025cipherbank}
Yu~Li, Qizhi Pei, Mengyuan Sun, Honglin Lin, Chenlin Ming, Xin Gao, Jiang Wu, Conghui He, and Lijun Wu.
\newblock {CipherBank}: Exploring the boundary of {LLM} reasoning capabilities through cryptography challenges.
\newblock \emph{arXiv preprint arXiv:2504.19093}, 2025.
\newblock URL \url{https://arxiv.org/abs/2504.19093}.

\bibitem[Lieber et~al.(2021)Lieber, Sharir, Lenz, and Shoham]{J1WhitePaper}
Opher Lieber, Or~Sharir, Barak Lenz, and Yoav Shoham.
\newblock Jurassic-1: Technical details and evaluation.
\newblock Technical report, AI21 Labs, August 2021.

\bibitem[Loshchilov and Hutter(2019)]{loshchilov2019decoupledweightdecayregularization}
Ilya Loshchilov and Frank Hutter.
\newblock Decoupled weight decay regularization, 2019.
\newblock URL \url{https://arxiv.org/abs/1711.05101}.

\bibitem[Park et~al.(2023)Park, Kim, and Moon]{parkcryptography7030035}
Seonghwan Park, Hyunil Kim, and Inkyu Moon.
\newblock Automated classical cipher emulation attacks via unified unsupervised generative adversarial networks.
\newblock \emph{Cryptography}, 7\penalty0 (3), 2023.
\newblock ISSN 2410-387X.
\newblock \doi{10.3390/cryptography7030035}.
\newblock URL \url{https://www.mdpi.com/2410-387X/7/3/35}.

\bibitem[Radford et~al.(2019)Radford, Wu, Child, Luan, Amodei, and Sutskever]{radford2019language}
Alec Radford, Jeff Wu, Rewon Child, David Luan, Dario Amodei, and Ilya Sutskever.
\newblock Language models are unsupervised multitask learners.
\newblock 2019.

\bibitem[Rae et~al.(2022)Rae, Borgeaud, Cai, Millican, Hoffmann, Song, Aslanides, Henderson, Ring, Young, Rutherford, Hennigan, Menick, Cassirer, Powell, van~den Driessche, Hendricks, Rauh, Huang, Glaese, Welbl, Dathathri, Huang, Uesato, Mellor, Higgins, Creswell, McAleese, Wu, Elsen, Jayakumar, Buchatskaya, Budden, Sutherland, Simonyan, Paganini, Sifre, Martens, Li, Kuncoro, Nematzadeh, Gribovskaya, Donato, Lazaridou, Mensch, Lespiau, Tsimpoukelli, Grigorev, Fritz, Sottiaux, Pajarskas, Pohlen, Gong, Toyama, de~Masson~d'Autume, Li, Terzi, Mikulik, Babuschkin, Clark, de~Las~Casas, Guy, Jones, Bradbury, Johnson, Hechtman, Weidinger, Gabriel, Isaac, Lockhart, Osindero, Rimell, Dyer, Vinyals, Ayoub, Stanway, Bennett, Hassabis, Kavukcuoglu, and Irving]{rae2022scalinglanguagemodelsmethods}
Jack~W. Rae, Sebastian Borgeaud, Trevor Cai, Katie Millican, Jordan Hoffmann, Francis Song, John Aslanides, Sarah Henderson, Roman Ring, Susannah Young, Eliza Rutherford, Tom Hennigan, Jacob Menick, Albin Cassirer, Richard Powell, George van~den Driessche, Lisa~Anne Hendricks, Maribeth Rauh, Po-Sen Huang, Amelia Glaese, Johannes Welbl, Sumanth Dathathri, Saffron Huang, Jonathan Uesato, John Mellor, Irina Higgins, Antonia Creswell, Nat McAleese, Amy Wu, Erich Elsen, Siddhant Jayakumar, Elena Buchatskaya, David Budden, Esme Sutherland, Karen Simonyan, Michela Paganini, Laurent Sifre, Lena Martens, Xiang~Lorraine Li, Adhiguna Kuncoro, Aida Nematzadeh, Elena Gribovskaya, Domenic Donato, Angeliki Lazaridou, Arthur Mensch, Jean-Baptiste Lespiau, Maria Tsimpoukelli, Nikolai Grigorev, Doug Fritz, Thibault Sottiaux, Mantas Pajarskas, Toby Pohlen, Zhitao Gong, Daniel Toyama, Cyprien de~Masson~d'Autume, Yujia Li, Tayfun Terzi, Vladimir Mikulik, Igor Babuschkin, Aidan Clark, Diego de~Las~Casas, Aurelia Guy, Chris Jones,
  James Bradbury, Matthew Johnson, Blake Hechtman, Laura Weidinger, Iason Gabriel, William Isaac, Ed~Lockhart, Simon Osindero, Laura Rimell, Chris Dyer, Oriol Vinyals, Kareem Ayoub, Jeff Stanway, Lorrayne Bennett, Demis Hassabis, Koray Kavukcuoglu, and Geoffrey Irving.
\newblock Scaling language models: Methods, analysis \& insights from training gopher, 2022.
\newblock URL \url{https://arxiv.org/abs/2112.11446}.

\bibitem[Reuter et~al.(2025)Reuter, Rudner, Fortuin, and R{\"u}gamer]{reuter2025can}
Arik Reuter, Tim G.~J. Rudner, Vincent Fortuin, and David R{\"u}gamer.
\newblock Can transformers learn full bayesian inference in context?
\newblock \emph{arXiv preprint arXiv:2501.16825}, 2025.
\newblock URL \url{https://arxiv.org/abs/2501.16825}.

\bibitem[Wang et~al.(2025)Wang, Wang, Ying, and Wang]{wang2025can}
Qixun Wang, Yifei Wang, Xianghua Ying, and Yisen Wang.
\newblock Can in-context learning really generalize to out-of-distribution tasks?
\newblock In \emph{International Conference on Learning Representations}, 2025.
\newblock URL \url{https://openreview.net/forum?id=INe4otjryz}.

\bibitem[Wenger et~al.(2023)Wenger, Bubeck, Garg, et~al.]{wenger2023salsa}
Ethan Wenger, Sébastien Bubeck, Shyam Garg, et~al.
\newblock Salsa: Solving lattice problems with large language models.
\newblock arXiv:2311.01234, 2023.
\newblock URL \url{https://arxiv.org/abs/2311.01234}.

\bibitem[Yang et~al.(2024)Yang, Lee, Nowak, and Papailiopoulos]{yang2024looped}
Liu Yang, Kangwook Lee, Robert Nowak, and Dimitris Papailiopoulos.
\newblock Looped transformers are better at learning learning algorithms.
\newblock In \emph{International Conference on Learning Representations}, 2024.
\newblock URL \url{https://arxiv.org/abs/2311.12424}.

\end{thebibliography}


\newpage

\appendix

\section{Training details}
\label{sec:trainingdetails}
We use a batch size of $64$ and train for $20$k total steps. The learning rate is set to $10^{-3}$, with a weight decay of $0.1$, but no learning rate scheduler. Furthermore, we employ the AdamW optimizer \citep{loshchilov2019decoupledweightdecayregularization} over Adam for its decoupled weight decaying, which has been shown to improve generalization and training stability. All training and experiments are done on a single NVIDIA RTX A5000 GPU.

\section{Ablation}
\label{sec:ablations}

\subsection{Batch Size for Mono-Alphabetic Substitution Ciphers}
\begin{figure}[htbp]
    \centering
    \begin{subfigure}[b]{1\linewidth}
        \centering
        \includegraphics[width=\linewidth]{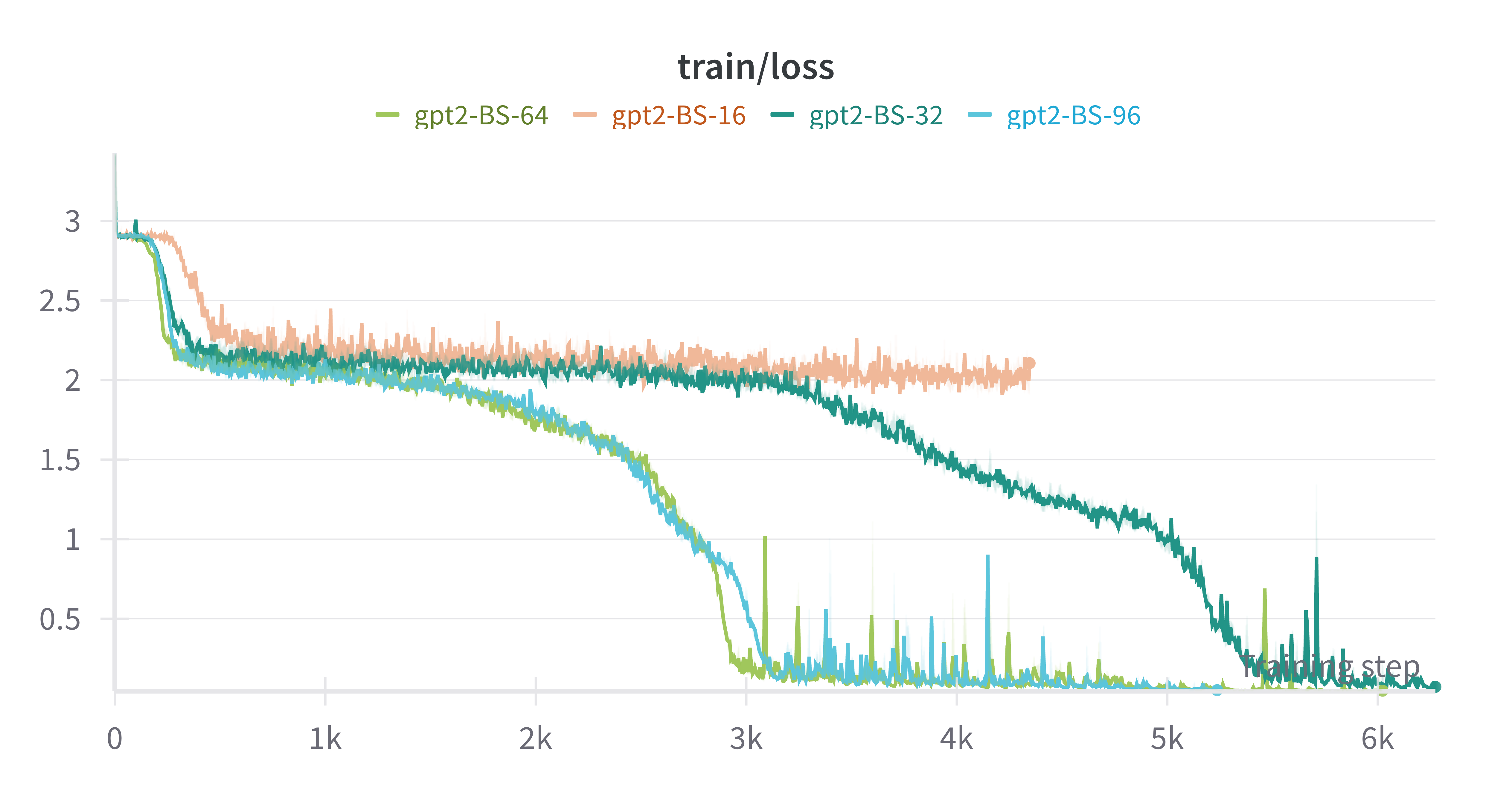}
        \caption{Represents our training loss for ICL for variable-length Vigenère ciphers. The batch size 16 seems to stabilize at loss value 2, while the other batch sizes for training at size 64 and 96 stabilize at around the same value of 0, with 32 taking longer to converge to $0$ loss}
        \label{fig:subfig1}
    \end{subfigure}
    \hfill
    \begin{subfigure}[b]{1\linewidth}
        \centering
        \includegraphics[width=\linewidth]{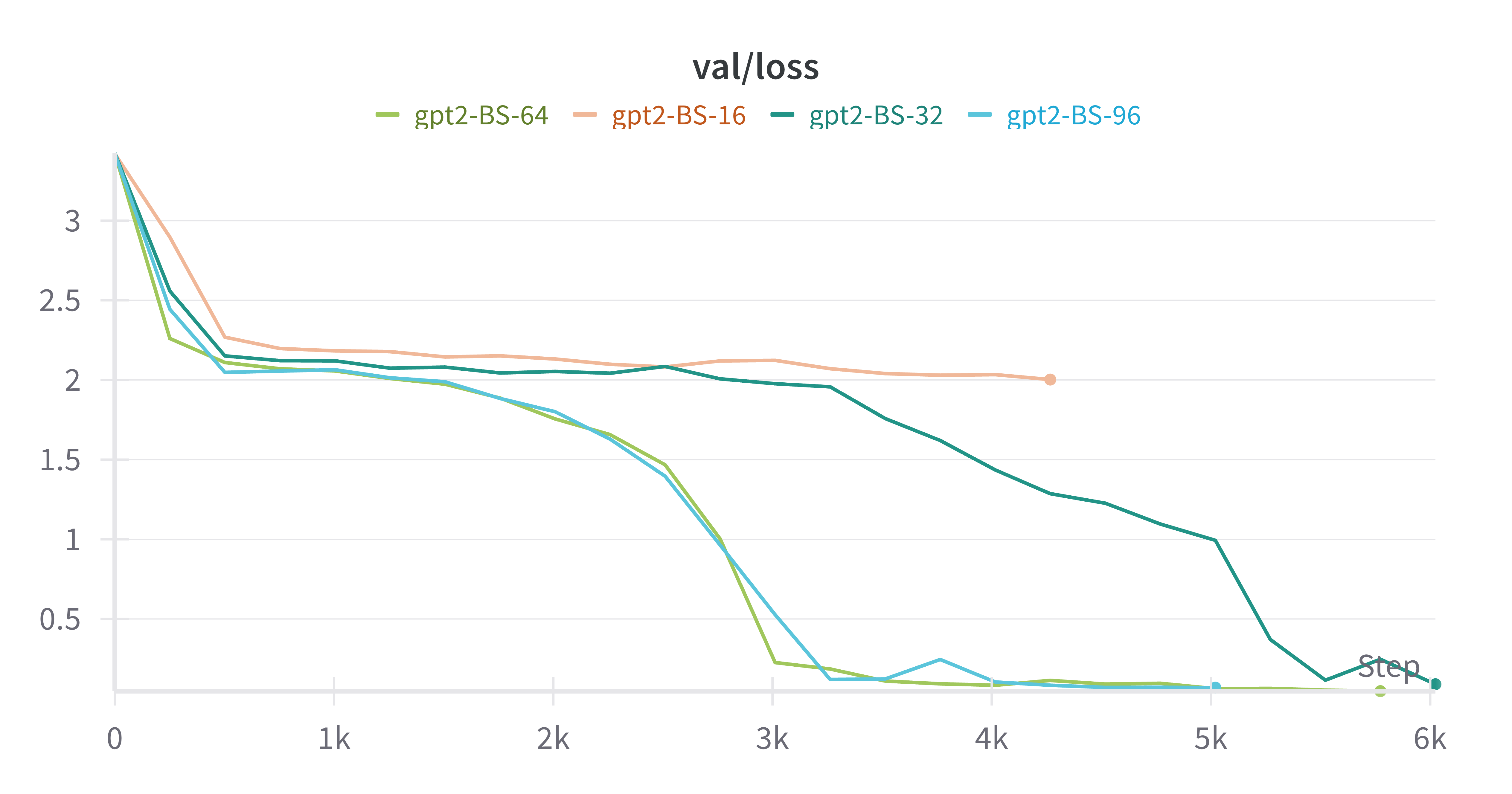}
        \caption{Validation losses perform very similar to the training losses, which means the model generalizes well on our testing data.}
        \label{fig:subfig2}
    \end{subfigure}
    \caption{We consider different batch sizes. Namely, batch sizes 16, 32, 64, and 96.}
    \label{fig:main-fig}
\end{figure}

\subsection{Context Length for Mono-Alphabetic Substitution Ciphers}
\begin{figure}[htbp]
    \centering
    \begin{subfigure}[b]{1\linewidth}
        \centering
        \includegraphics[width=\linewidth]{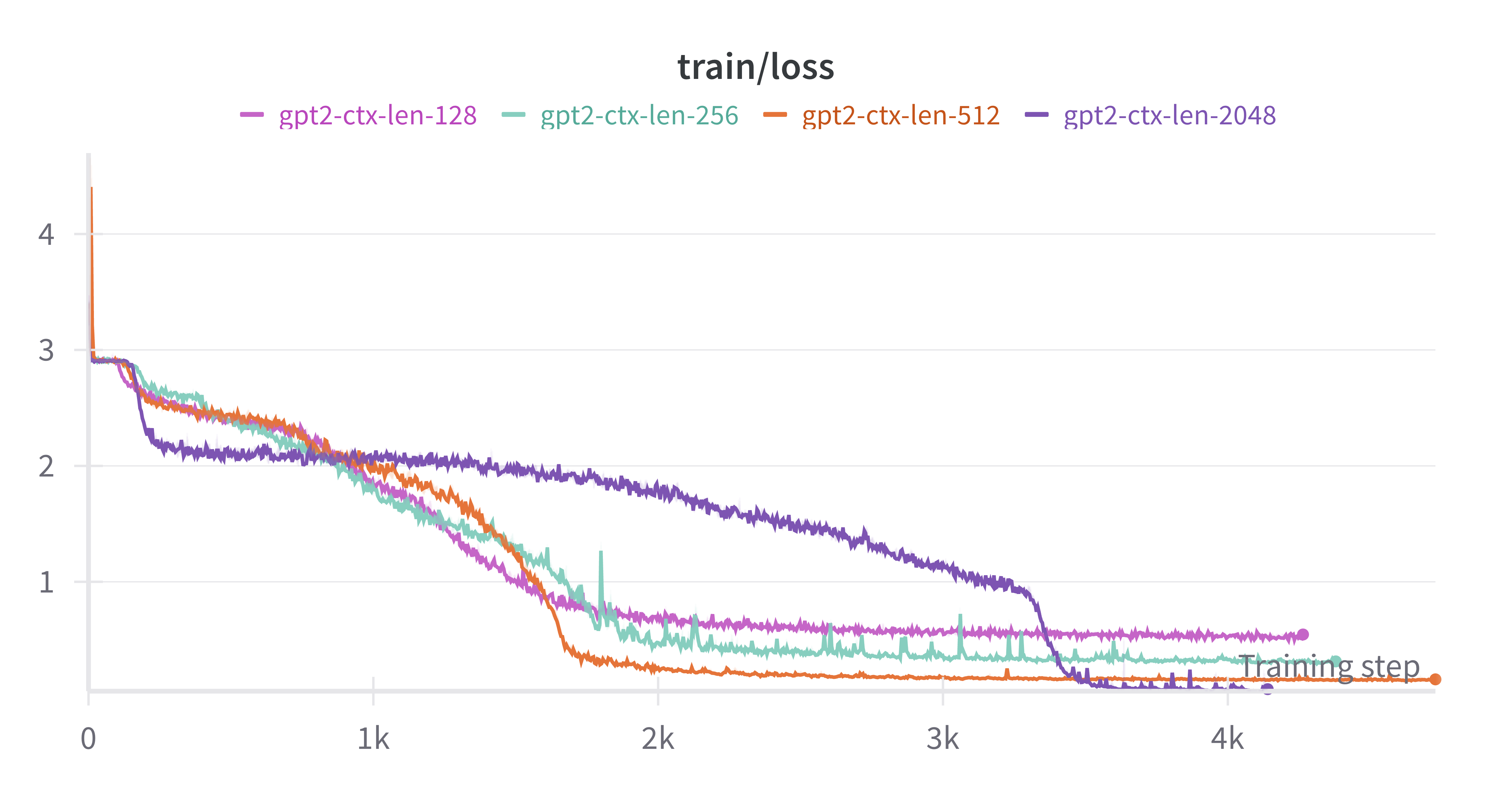}
        \caption{Graph above examines the affect of context length on the training of our trained GPT2 model. All context lengths seem to converge together, with context length 2048 being the most fluctuatous, with the other lengths converging similarly.}
        \label{fig:subfig1}
    \end{subfigure}
    \hfill
    \begin{subfigure}[b]{1\linewidth}
        \centering
        \includegraphics[width=\linewidth]{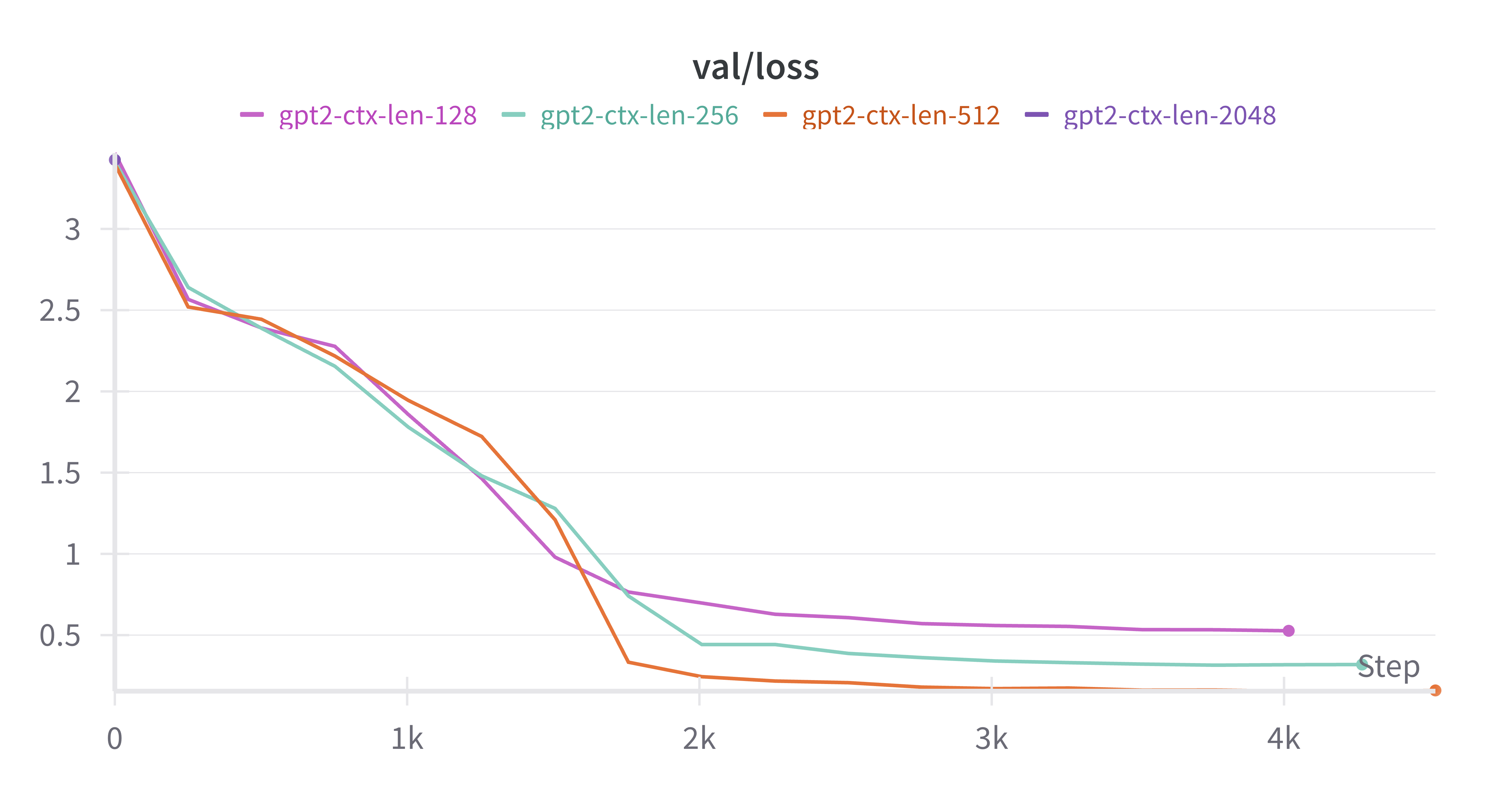}
        \caption{Validation losses are quite similar, where now context length 2048 converges more smoothly than the training loss, where the other context lengths converge similarly to the training loss. This implies that our model is trained and generalizes well.}
        \label{fig:subfig2}
    \end{subfigure}
    \caption{We consider different context lengths (with a fixed batch size of 64). Namely, batch sizes 128, 256, 512, and 2048.}
    \label{fig:main-fig}
\end{figure}

\end{document}